\documentclass[10pt,letterpaper]{article}

\usepackage{cogsci}
\usepackage{amsmath,amssymb}
\usepackage{commath}
\usepackage{booktabs}
\usepackage{algorithm}
\usepackage{algpseudocode}
\usepackage{tabularx}
\usepackage{caption}
\usepackage{subcaption}
\usepackage{graphicx}
\usepackage{amsmath,amssymb}
\usepackage{relsize}
\cogscifinalcopy 

\usepackage[
    backend=biber,
    style=apa,
    natbib=true,
    doi=false,
    isbn=false,
    url=false,
]{biblatex}
\usepackage{pslatex}
\usepackage{float} 

\usepackage{graphicx}
\graphicspath{ {./images/} }
\newcommand\blfootnote[1]{%
  \begingroup
  \renewcommand\thefootnote{}\footnote{#1}%
  \addtocounter{footnote}{-1}%
  \endgroup
}

\title{What's in the Box? Reasoning about Unseen Objects from Multimodal Cues}

\author{\\\\{\large \bf Lance Ying$^{*1,2}$, Daniel Xu$^{*1}$, Alicia Zhang$^{1}$, Katherine M. Collins$^{1}$, Max H. Siegel$^{1}$, Joshua B. Tenenbaum$^{1}$} \\ \\
  $^{1}$Massachusetts Institute of Technology, Cambridge, MA, USA \\
  $^{2}$Harvard University, Cambridge, MA, USA\\
  $^{*}$Equal Contribution
  }

\begin{document}

\maketitle

\begin{abstract}

People regularly make inferences about objects in the world that they cannot see by flexibly integrating information from multiple sources: auditory and visual cues, language, and our prior beliefs and knowledge about the scene. How are we able to so flexibly integrate many sources of information to make sense of the world around us, even if we have no direct knowledge? In this work, we propose a neurosymbolic model that uses neural networks to parse open-ended multimodal inputs and then applies a Bayesian model to integrate different sources of information to evaluate different hypotheses. We evaluate our model with a novel object guessing game called ``What's in the Box?'' where humans and models watch a video clip of an experimenter shaking boxes and then try to guess the objects inside the boxes. Through a human experiment, we show that our model correlates strongly with human judgments, whereas unimodal ablated models and large multimodal neural model baselines showed poor correlation.

\end{abstract}

\section{Introduction}

Reasoning about the presence and properties of occluded objects is a fundamental aspect of human intelligence, crucial for navigating and interacting with a complex world. This ability is based on the integration of various sensory cues and contextual information to form probabilistic judgments. Consider the following scenario. You are locating your package in a cluttered mail room. Which box is yours? Rather than exhaustively examining each package, you can efficiently leverage multimodal cues such as size, weight, and textual labels to rapidly narrow the search space and infer the likely location of your target. This capacity to infer hidden states based on incomplete and potentially ambiguous observations raises a critical question: how do humans effectively integrate multimodal information to reason about the unobserved?\blfootnote{Published as a conference paper at CogSci 2025}

\begin{figure}[t!]
\centering
\includegraphics[width=0.45\textwidth]{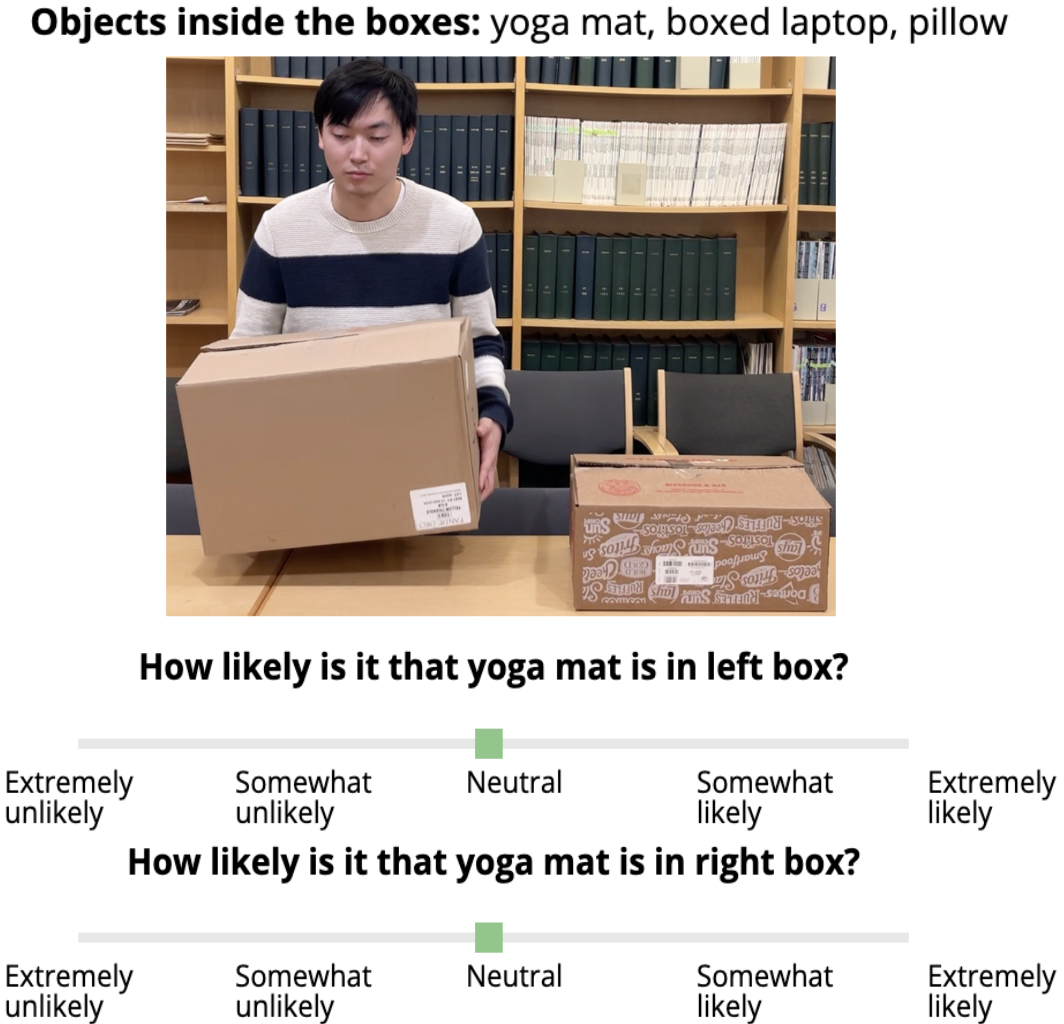}
\caption{The What's in the Box (WiTB) game. In this game, the participants are given a written list of objects hidden in boxes. They then watch a video of a human experimenter shaking the box. The participants are then asked to guess where each object is hidden based on the visual and audio cues.}
\label{fig: interface}
\end{figure}

\begin{figure*}[t!]
\includegraphics[width=1\textwidth]{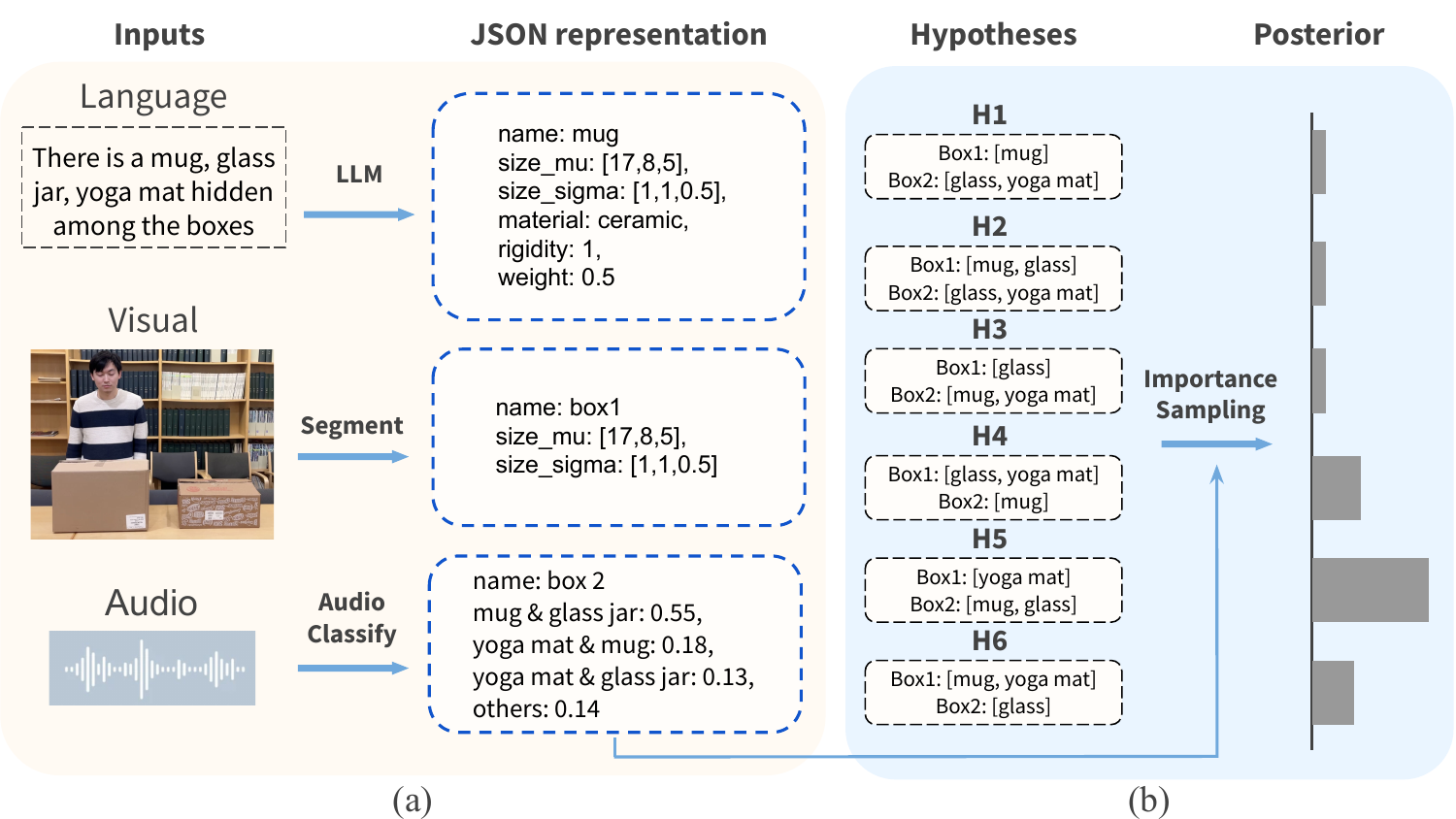}
\caption{Our neurosymbolic model. (a) The model first uses neural networks to parse multimodal input to a structured JSON representation. (b) For a set of objects and boxes, the model generate all hypotheses of object placements among boxes. Then the hypotheses are evaluated based on the visual and audio information to generate a posterior distribution over the hypotheses, which can be marginalized to infer object placements.}
\label{fig: main}
\end{figure*}

Accounting for uncertainty regarding unobserved phenomena is a core aspect of human intelligence and has been extensively studied in cognitive science. Research on object permanence in developmental psychology \citep{piaget1954construction} highlights the early emergence of the ability to represent and reason about objects that are no longer directly perceived. Classic paradigms such as the Wason selection task \citep{wason1968reasoning} have explored the use of deductive reasoning to answer questions about unseen objects. More recent work has focused on how humans infer hidden properties of objects based on partial observations, incorporating probabilistic models of scene understanding and physical reasoning. These models commonly use a physics engine as a generative model of visual percepts, which can then be inverted to infer the physical attributes of the underlying hidden object \citep{battaglia2013simulation,lake2017building, yildirim2016perceiving}. However, these computational models typically focus on vision as the single modality of perceptual cue and evaluate the model in well-controlled simulated environments. The models often require extensive hand-engineering to restrict the space of the object's physical attributes for running the simulation in order to make inference tractable. Consequently, these existing cognitive models may not be able to account for the complexity of real-world scenarios that require integrating cues from multiple modalities, and the physical properties of the candidate objects are not given a priori.

On the other hand, a complementary line of research in cognitive science has extensively documented the integration of multimodal cues in low-level perception. A typical setting presents observers with a multimodal stimulus (e.g. a flash of light and a noise burst), each modality giving a noisy estimate of a scene quantity (e.g. source location) \citep{alais2004ventriloquist}). This paradigm is motivated by a canonical multisensory integration model, which applies Bayes' rule to derive an optimal estimate by combining information from each sense \citep{trommershauser2011sensory,ernst2007humans,kording2007causal}. Human experiments have found remarkable agreement between the predictions of this model - which performs a weighted average of the modality-specific estimates - and human judgments, with notable examples in audiovisual \citep{alais2004ventriloquist, battaglia2003bayesian} and visual-haptic  \citep{ernst2007humans} processing. However, multimodal \textit{reasoning}, potentially incorporating perceptual information as well, is far less studied in cognitive science.



Rapid progress has also been made in multimodal reasoning from the artificial intelligence (AI) and robotics community, particularly through deep learning over massive datasets \citep{nam2017dual}. The advent of versatile Vision-Language Models (VLMs) and Large Language Models (LLMs) has further expanded these capabilities, allowing for generalization to previously unseen scenarios \citep{wang2024exploring, ahn2022can}. However, critical questions remain as to how well these models truly grasp physical and visual reasoning. While they excel at pattern recognition and language understanding, there are ongoing debates about their capacity for scene understanding, multimodal reasoning, and interpreting the causal relationships inherent in the physical world. 

In this paper, we present a new neurosymbolic model designed to perform robust reasoning about hidden objects from complex and ambiguous multimodal inputs. Leveraging a suite of state-of-the-art neural networks for processing text, audio, and visual data, our model constructs a formal representation of the observed scene. Subsequently, a Bayesian inference engine updates hypotheses about the hidden objects based on these observations. Such a neurosymbolic structure combines the strength of both data-driven large neural models and a Bayesian architecture for integrating cues from different modalities for robust reasoning about unseen objects. 

We evaluate our model on a novel object guessing task that we call ``What's in the Box?'' (WiTB), wherein objects are concealed within boxes, and an observer must infer their contents by analyzing a human participant's interactions with the boxes, including lifting and shaking. We demonstrate that the proposed neurosymbolic model effectively integrates visual, textual, and auditory information to achieve human-like performance in reasoning about object placements. Critically, unimodal models exhibit significantly poorer performance, highlighting the crucial role of multimodal integration in this task. This work takes one step towards helping us better understand how humans can flexibly and reliably infer information about objects we cannot see from diverse information sources.

\section{Computational Model}

Our model is shown in Fig. \ref{fig: main}. Similar to prior work on neurosymbolic reasoning \citep{ying2023neuro,wong2023word,hsu2023ns3d}, our neurosymbolic model consists of two modules: (1) a neural module translates multimodal inputs into formal representation, and (2) an engine for probabilistic inference to form the final graded judgment. That is: in the second module, our model performs probabilistic inference over the parsed symbolic representation to derive the posterior distribution over the object placements.

\subsection{Parsing and representing multimodal inputs}

Reasoning about multimodal scenes is often complex as it requires integrating information across different modalities. Following prior work on scene and semantic parsing using large foundation models \citep{ying2025understanding, liu2015semantic}, we use a variety of neural models to parse multimodal inputs into structured symbolic forms. 

\paragraph{Language:} The linguistic information provided to an observer includes names of the objects hidden among the boxes. However, human language is often abstract and ambiguous. How do we know about the properties of the hidden objects, such as a pillow, without seeing them? Humans often rely on their knowledge and memory (past observations) as clues. In our model, we prompt a state-of-the-art large language model (LLM) to generate attributes of the unseen object, including its geometric dimensions, weights, materials, and rigidity (The degree to which the object can be compressed or folded in any dimension). This is because the LLMs, trained on a large amount of real-world data, have likely encountered more objects than any person and can provide reasonable guesses about them. Our model uses the Llama 3.1 70B model as the LLM parser.

Furthermore, to capture the uncertainty about the objects from the language input (e.g. pillows can have various sizes), we prompted the LLM to output standard deviations for some key attributes, such as the physical dimensions, from which we can model uncertainty by assuming a normal distribution over these variables.

\paragraph{Vision:} From the visual inputs, we can estimate the size of the boxes present in the scene. We prompt Gemini 2.0 Flash (Gemini \cite{geminiteam2024geminifamilyhighlycapable}) with the first frame of the video and ask the model to return the dimensions of the boxes in the format specified in Figure \ref{fig: main}.

\paragraph{Audio:} The sound made by shaking the boxes can also provide us with clues on what's inside them. In our model, we use an audio classification algorithm CLAP (\cite{elizalde2022clap}) to generate a probability distribution over the type of the sounds from the audio track with object names as candidate labels. This allows us to calculate the posterior probability of objects in any box conditioned on the audio of the box when shaken by a human.

\subsection{Generating and evaluating hypotheses}

To infer the placement of objects among boxes, we adopt an approach inspired by particle filtering \citep{wills2023sequential}. We first initialize all the hypotheses $H = \{H^1...H^n\}$, each representing a unique way of placing objects in distinct objects. With N objects in K boxes, this would generate $\abs{H}=K!S(N, K)$ possible placements, represented as an ordered set of lists of objects, Here $S(N, K)$ is the Stirling numbers of the second kind. The observer has a uniform prior belief about the placement $b_0 = P(H)$. Then the observer performs a belief update conditioned on the observed multimodal inputs.

We denote hypothesis $H^i = H^i_1,... H^i_n$ where $H^i_n$ is the set of items in box n according to hypothesis $H^i$. We then denote the audio observation as $A = A_1,...,A_n$, visual observation as $O = O_1, O_2,..., O_n$ where $A_n$ is the audio of box n. Since we assume a uniform prior, this is 

\begin{align}
    P(H| O,A) &\propto P(O,A|H) = P(O|H) P(A|H) \\ \nonumber
    &\propto P(O|H) P(H|A) = \prod_i P(O_i|H_i)P(H_i|A_i) 
\end{align}

Here, since the audio and visual signals are both ambiguous and their underlying joint probability distribution is often not accessible, we treat them as conditionally independent as a reasonable approximation. We also manipulate the conditional probabilities to compute $P(H_i|A_i)$ because the audio likelihood function $P(A_i|H_i)$, which requires a generative model for audio, is difficult to estimate while state-of-the-art audio classification models can readily output posterior distributions $P(H_i|A_i)$.

For evaluating $P(O_i|H_i^n)$, we use rejection sampling by checking whether the set of items $H_i^n$ can fit in the box $i$. To account for uncertainties about the physical attributes of the boxes and objects, we sample their dimensions under a normal distribution and apply rejection sampling 1000 times to produce a continuous probability distribution.

We then evaluate $P(H_i^n|A_i)$ by querying the CLAP model with the audio segment $A_i$ and item labels which are all possible items in the scenario.

\begin{align}
    P(H_i^n|A_i) = \prod_{o \in H_i^n} P(o|A_i)
\end{align}

\begin{figure*}[t!]
\includegraphics[width=1\textwidth]{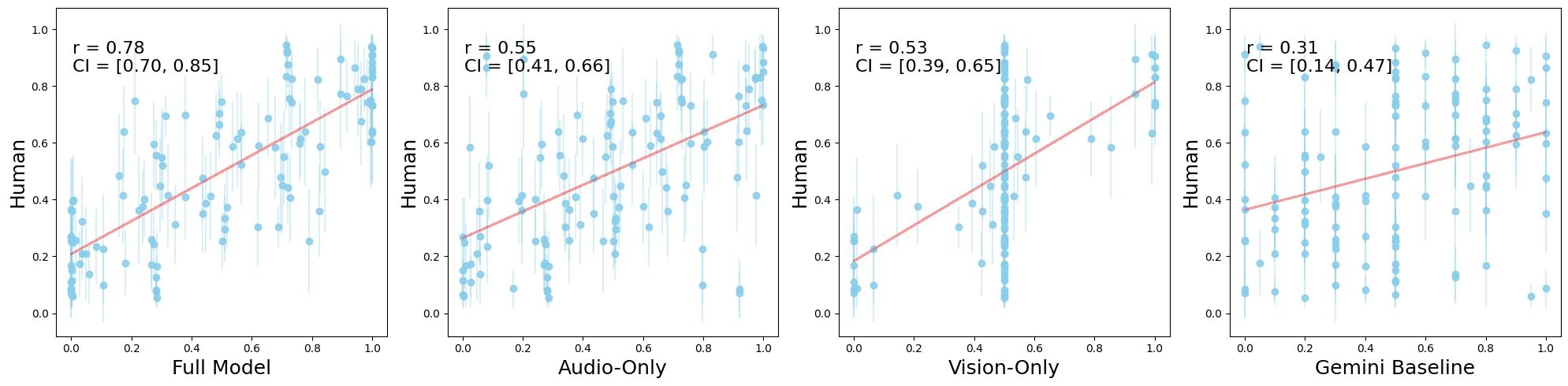}
\caption{Correlation plots comparing belief judgments from humans (y-axis) against models (x-axis). Each dot represents a probability rating on a scale from 1 to 100 (e.g. how likely is it that object X is in the left box). Error bars show standard error and CI indicates 95\% confidence interval. Our full model shows a significantly better fit to human judgment than ablated unimodal baselines and the Gemini model. Error bars indicate standard error.}
\label{fig: scatter}
\end{figure*}

Once the model computes the posterior distributions over the hypotheses, it then marginalize over all hypotheses to compute the distribution for any individual object.

\vspace{1cm}




\begin{figure*}[h!]
\centering
    \begin{subfigure}{.45\textwidth}
  \centering
  \includegraphics[width=.96\linewidth]{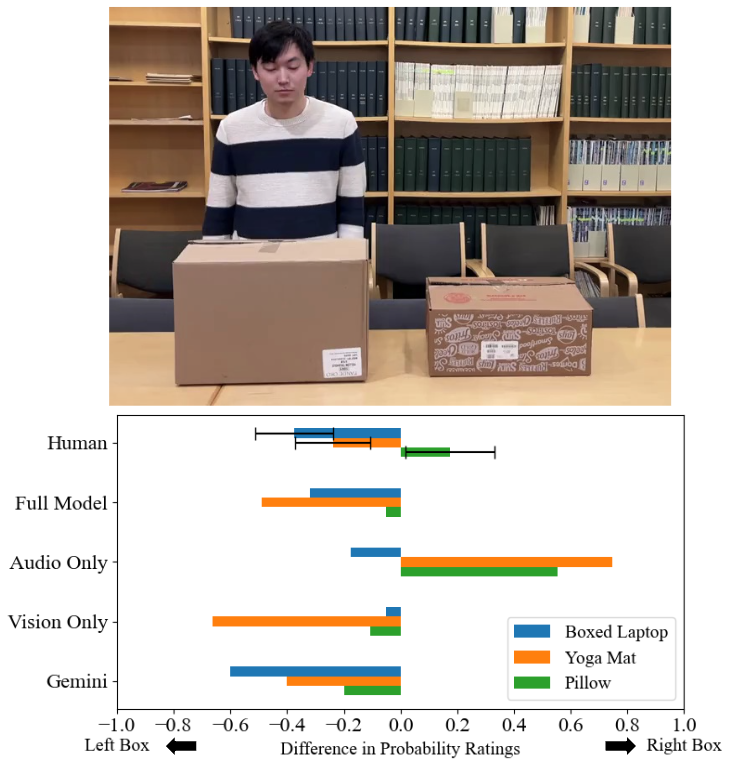}
  \caption{Scenario A: a yoga mat, a laptop packaged inside a box, and a pillow hidden inside two boxes. \\ Video link: https://youtu.be/JE4ggHKfRss}
  \label{fig:bar1}
\end{subfigure}%
\hspace{5mm}
\begin{subfigure}{.45\textwidth}
  \centering
  \includegraphics[width=.96\linewidth]{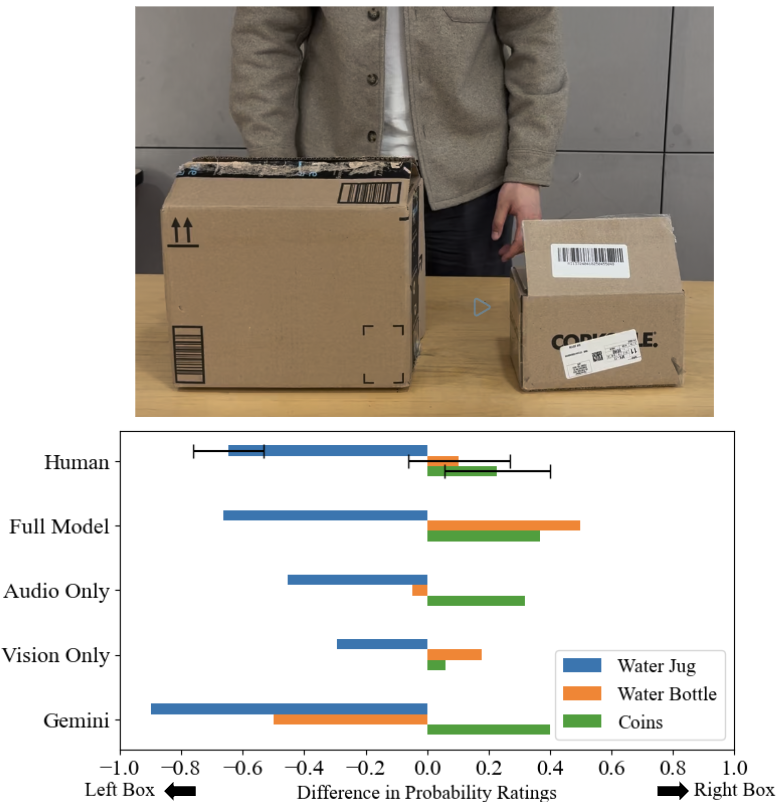}
  \caption{Scenario B: water jug, water bottle, and coins hidden inside two boxes. \\ Video link: https://youtu.be/TdZHEkuGDgM}
  \label{fig:bar2}
\end{subfigure}
\caption{Two qualitative examples comparing model and human ratings on the location of the objects. The bars represent the averaged human or model differences in the probability rating of an item being inside the left versus right box. Error bars indicate standard error.}
\label{fig:qualitative}
\end{figure*}
\section{Experiment}

\paragraph{Domain and Scenarios}
We constructed 25 stimuli, each with 2 to 5 household objects, such as water bottles, mugs, plates, laptops, etc., hidden inside 2 boxes. The items were carefully chosen to represent various shapes, sizes and materials. The type of materials cover ceramics, metals, plastic, wood, etc., which produce a variety of sounds in the video clip. The boxes also vary in sizes, where some boxes can hold all objects and some may hold one or two small objects. In each stimulus, we record a 2 to 3 seconds video of a human experimenter shaking the boxes. We excluded 5 stimuli in the experiment due to low agreement among human participants (split-half correlation less than 0.8).

\paragraph{Human Participants}
We recruited 54 participants over Prolific (mean age = 37; 29 male, 24 female, 1 other). The experiment took place over a customized web interface, which is shown in Fig. \ref{fig: interface}. During the experiment, each participant is shown a video with sound and asked to evaluate the likelihood of the object hidden inside each box with a set of continuous dependent scales from 1 to 100 where the of scale for each item automatically sums to 100 across the box options as the user drags.

\paragraph{Baselines:}
Our model has two critical components: reasoning about the objects from both visual and auditory cues, and then combining different sensory inputs to perform belief updates. To evaluate the criticality of the integration of multimodal inputs, we consider two alternative models involving unimodal ablations, wherein we remove one of the sensory inputs.  

The Audio-Only model only receives auditory information as input. It uses the CLAP model to assign probability ratings for the object being inside any box given the sound, and then normalize the ratings over all boxes. In other words, the probability of object $o$ inside box $i$ would be 
\begin{align}
    \frac{P(o \in box_i |A_i)} {\sum_j  P(o \in box_j |A_j)}
\end{align}

On the other hand, the Vision-Only Model only receives visual information. Similar to the visual module in the full model, the Vision-only model uses geometric properties to guess where the object may be hidden (e.g. a yoga mat is more likely to be in a big box than a small one).

Additionally, we evaluate a state-of-the-art vision-language foundation model, Gemini 2.0 Flash, as a neural baseline. The VLM model was given a video with audio and provided the same instructions as human participants to evaluate the probability distributions of hidden objects across boxes.

\section{Results}

\paragraph{Quantitative Analysis:} As shown in Figure \ref{fig: scatter}, the Full Model correlated strongly with human judgment, with $r = 0.78$, while the ablated Audio-Only Model and Vision-Only Model performed worse, with $r = 0.55$ and $ r = 0.52$, respectively. On the other hand, we also find the VLM model had a low correlation of $r = 0.31$ against human judgments, indicating that large foundation models still cannot reliably reason about ambiguous multimodal inputs in a human-like way.

Taken together, these results showcase the promise of a Bayesian approach integrating different modalities of inputs to reason about objects in ambiguous and highly uncertain scenarios.

\paragraph{Qualitative Analysis}
We highlight two examples for qualitative analysis comparing the model performances. The visual layouts for the two examples are shown in Fig. \ref{fig:qualitative}. 

In Scenario A, based on the visual information, the model is confident that the yoga mat is inside the left box because it is unlikely to fit inside the right box. However, the vision-only model is uncertain about the location of the laptop and the pillow. The audio model, on the other hand, finds that the laptop is more likely to be inside box 1 due to the collision sound it makes. Combining these two sources, the full model is able to make a graded judgment about the location of these objects that mirrors human judgment, whereas audio-only and vision-only models made different judgments based on unimodal information. On the other hand, the Gemini model believes all three objects are likely to be inside box 1, which reflects poor physical reasoning skills.

In Scenario B, we show a scenario where a water jug, a water bottle, and coins are distributed between two boxes of differing sizes. Based on the visual information on box sizes, the model finds the water jug is more likely to be inside box 1 since it might not fit inside box 2. However, it is uncertain where the coins are because they are small and could fit in either box. The audio model is able to determine where the coins are because of the distinct jingling sounds, whereas the water jug and bottle make the same sound and therefore cannot be distinguished by the audio model. Combining these two sources, the full model is able to make a judgment that mirrors human judgment, where neither ablated model would be able to if it had been restricted to only one modality. In contrast, the Gemini model can approximate a subset of the objects, but seems to be unable to reason about second-order effects of an object placement. 

Interestingly, in these examples, we find that the resulting probability judgments of the full model is not simply an average over audio and visual model outputs, as the joint inference over object placements conditioned on audio and visual information is performed over all hypotheses before marginalized for each individual object ratings.


\paragraph{Error Analysis:} 
We observe that in a few scenarios, our model is quite uncertain (almost equally likely in any of the two boxes) while the humans are more confident in their judgments on where the item is located. One possibility is that the visual cues our models are using are still limited, whereas humans may be leveraging more kinds of visual information to reason about the object placements. For instance, humans are able to infer the weight and size of the items inside the box based on the motion of the box, which they can use to update beliefs about the box's content.

Additionally, we noticed that the audio model sometimes failed to pick up nuanced audio information when multiple sounds were present. For instance, the model may not pick up plastic sound when it's mixed with metallic sound, whereas humans are comparably better at recognizing and parsing subtle audio cues.

\section{Discussion and Future Directions}

In this paper, we introduce a neurosymbolic model that performs robust yet generalizable reasoning based on multimodal cues. The model uses state-of-the-art neural models as the perception module and then uses a Bayesian model for initializing and updating beliefs on hypotheses about the unseen objects. We evaluate the model on a novel paradigm called ``What's in the Box?'' (WiTB), wherein models and humans watch experimenters interact with boxes and guess which items are hidden in which box. Our results show that the proposed neurosymbolic model correlates strongly with human judgments, whereas other ablated models and state-of-the-art neural vision baselines perform poorly.

Our model offers significant contributions to both cognitive science and artificial intelligence. By integrating the pattern recognition capabilities of neural networks with the structured reasoning of Bayesian models, we provide cognitive scientists with a new tool to investigate human inference processes in more open-ended settings, particularly under conditions of uncertainty and with complex multimodal information. This neurosymbolic architecture also holds promise for the development of more intelligent robots, enabling them to reason about the physical properties of unseen objects by effectively combining diverse sensory cues, thus approaching human-like reasoning and scene understanding.

There are, however, a few important limitations and open directions for improvement. Firstly, our current model assumes that all sources of information are equally weighted. In reality, humans likely adaptively weigh different cues based on their reliability and relevance to the task \cite{jacobs2002determines, schertz2020phonetic}. For instance, a distinct sound might overshadow a partially occluded visual cue. Future iterations of the model should explore mechanisms for learning and dynamically adjusting the weights associated with each modality. 

Additionally, future studies can expand the kinds of visual cues we considered in our model. For example, our model currently does not infer the weight of the boxes, which can be informative to what objects may be inside. Writings on the box (e.g. an IKEA box) can also be used to infer the kinds of objects it contains.

The auditory component of our current model can also be improved. The lightweight audio model used in our model is less sensitive to ambiguous and low-volume sounds, which sometimes fails to use nuanced audio cues to reason about unseen objects. Future work can explore more sophisticated audio models, especially the ones trained on large-scale sound datasets, to improve the performance.

For next steps, we also plan on extending the WiTB paradigm to more open-ended settings. Rather than answering questions about objects from a pre-defined list, we can query the model and humans to guess what objects are in the box in an open-ended way based on multimodal cues and compare the distribution of answers given by humans and models, as in \cite{ying2025benchmarking}. This could allow us to study and capture the richness of humans' perception of what's out there in the world that we cannot directly observe in an open-ended environment.


\printbibliography

\end{document}